\documentclass{article}

\usepackage{PRIMEarxiv}

\usepackage[utf8]{inputenc} 
\usepackage[T1]{fontenc}    
\usepackage{hyperref}       
\usepackage{url}            
\usepackage{booktabs}       
\usepackage{amsfonts}       
\usepackage{nicefrac}       
\usepackage{microtype}      
\usepackage{lipsum}
\usepackage{fancyhdr}       
\usepackage{graphicx}       
\graphicspath{{media/}}     
\usepackage{algpseudocode}
\usepackage{algorithm}
\usepackage{amsmath, amssymb, amsthm}
\usepackage{enumitem}
\setlist[description]{leftmargin=1cm, labelindent=1cm}

\title{A Self-Adaptive Genetic Algorithm for the Flying Sidekick Travelling Salesman Problem
}

\author{
  Ted Pilcher \\
  University of Nottingham \\
  \texttt{psytp3@nottingham.ac.uk} \\
  \today \\
}

\begin{document}
\maketitle

\begin{abstract}
This paper presents a novel approach to solving the Flying Sidekick Travelling Salesman Problem (FSTSP) using a state-of-the-art self-adaptive genetic algorithm. The Flying Sidekick Travelling Salesman Problem is a combinatorial optimisation problem that extends the Travelling Salesman Problem (TSP) by introducing the use of drones. In FSTSP, the objective is to minimise the total time to visit all locations while strategically deploying a drone to serve hard-to-reach customer locations. Also, to the best of my knowledge, this is the first time a self-adaptive genetic algorithm (GA) has been used to solve the FSTSP problem. Experimental results on smaller-sized problem instances demonstrate that this algorithm can find a higher quantity of optimal solutions and a lower percentage gap to the optimal solution compared to rival algorithms. Moreover, on larger-sized problem instances, this algorithm outperforms all rival algorithms on each problem size while maintaining a reasonably low computation time. 
\end{abstract}


\section{Introduction}
The e-commerce industry has become extremely competitive in recent years, driven by the increasing demand for efficient and rapid delivery services, like same-day and next-day delivery services provided by Amazon. Last-mile delivery, comprising all logistic activities related to the delivery of shipments to customer households in urban areas \cite{boysen2021last}, plays a pivotal role in meeting these demands, but faces challenges including high costs, time pressures and an ageing workforce, among other issues \cite{boysen2021last}. To address these challenges and meet the demand for rapid delivery services, the concept of using unmanned aerial vehicles (UAVs) or drones has gained prominence as a potential solution. Conglomerates in the industry such as Amazon and DHL have already embraced drone technology for parcel delivery. Their initiatives like Prime Air and DHL’s Parcelcopter showcase the advancements in drone capabilities to deliver shipments, including their speed and versatility.

Traditionally, only trucks have been used for delivery and transportation due to their long-distance range and large capacity, which them a reliable choice. However, there are also limitations associated with using trucks for transportation, including their substantial size, slow speeds, high costs per mile and requirement for infrastructure. These drawbacks become particularly prominent in urban environments characterised by traffic congestion and access restrictions where the cost per mile can rise even higher due to lower average speeds. In contrast, the advantages of drones are their exceptional speed and that they don't rely on traditional infrastructure, like roads and bridges. Nevertheless, drones also carry limitations, including their limited endurance and parcel capacity. Therefore, employing both vehicles for delivery operations has emerged as a promising solution. Providing the potential to enhance delivery efficiency by combining the high delivery capacity of trucks with the agility and speed of drones. Companies like Mercedes-Benz have introduced innovative concepts like the Vision Van, which merges several innovative technologies for last-mile delivery operations including coordinating drone-truck delivery systems. This has led to the introduction of new routing problems, one of which is the Flying Sidekick Travelling Salesman Problem (FSTSP), first introduced by \cite{murray2015flying}. This problem involves coordinating the delivery of traditional driver-operated trucks with drones launched from the truck to serve customers in remote or hard-to-reach locations.

FSTSP is a challenging combinatorial optimisation problem classified within the NP-hard category. Consequently, exact solutions based on integer linear programming (ILP) formulations face considerable difficulties when dealing with larger instances of the problem, as the computation time required to find exact solutions increases exponentially with the linear increase of customer locations. Consequently, the adoption of heuristics and metaheuristics is proposed as a promising approach to efficiently tackle larger-sized instances of FSTSP. The genetic algorithm (GA), a metaheuristic, is used in this paper due it its advantages over regular heuristics and trajectory-based metaheuristics. Primarily, their ability to maintain a broader diversity of solutions by utilising a population of solutions rather than a single solution. Thereby allowing them to explore a broader search space of solutions and therefore, preventing premature convergence to sub-optimal solutions.

The main contributions of this paper are the proposed self-adaptive genetic algorithm that uses a wide range of crossover and problem-tailored mutation operators along with a novel two-stage mutation process designed specifically for FSTSP, where both the tour sequence and node types can be modified in a single generation.

The remainder of this paper will continue as follows: in section \ref{literature}, we present the related literature to last-mile delivery. In section \ref{problem}, a formal definition of FSTSP is provided along with a detailed description of the problem. Moreover, this section includes the assumptions made for this problem. In section \ref{proposed algorithm}, the self-adaptive genetic algorithm is presented and a detailed breakdown of each component is provided. In section \ref{results}, the experimental results are provided. This includes the process of setting the hyper-parameter values and the results obtained on both smaller and larger-sized problem instances along with the corresponding computation times. Finally, section \ref{conclusion} concludes with final remarks and discussions.

\section{Related Literature}
\label{literature}

Following the introduction of the FSTSP problem by \cite{murray2015flying}, several alternative formulations of the drone-routing problem have emerged. Noteworthy among them is the TSP-D problem, presented by \cite{agatz2018optimization}, which has received extensive research in recent years. TSP-D differs from FSTSP by allowing drones to return to the truck at the same customer location at which they launched from the truck. This special case is not allowed in FSTSP. Given the inherent similarities, this section will explore existing literature surrounding both TSP-D and FSTSP. This includes how exact methods have been employed to solve problem instances, as well as the utilisation of heuristics and metaheuristics to provide solutions to larger-sized problem instances.

\subsection{Exact Methods}
Exact methods use integer linear programming (ILP) formulations to generate provably optimal solutions to both the TSP-D and FSTSP problem. However, the scope of which exact solutions are applicable is limited when dealing with larger-sized problem instances. This limitation arises due to the inherent computational complexity of both TSP-D and FSTSP, as they are NP-hard problems. \cite{vasquez2021exact} presented an exact solution for TSP-D using a mixed-integer programming formulation which was able to exploit the model's structure to create two stages, (i) selecting and sequencing a subset of customers served by a truck and (ii) planning where to execute drone direct dispatches from the truck to each of the remaining customer locations. \cite{roberti2021exact, bouman2018dynamic} proposed exact solutions to TSP-D using dynamic programming to solve larger-sized instances optimally. Exact solutions using branch and cut algorithms were presented by \cite{boccia2021column, schermer2020b} for the FSTSP and TSP-D problems respectively. \cite{boccia2021column} combined the branch and cut algorithm with a column generation procedure for truck and drone synchronisation. \cite{dell2021algorithms} proposed an exact solution using a branch and bound algorithm for FSTSP with up to 15 customer locations. For larger instances, \cite{dell2021algorithms} also proposed a branch and bound algorithm with a heuristic to find high-quality solutions.

\subsection{Heuristics}
Due to the inherent limitations of exact solutions when dealing with larger problem instances, heuristics have also been utilised. For TSP-D, \cite{agatz2018optimization} presented route-first, cluster-second heuristics based on local search and dynamic programming. \cite{ha2018min} proposed a new variant of TSP-D to minimise the operational costs including synchronisation time and transportation costs. Moreover, \cite{ha2018min} proposed two procedures, (i) the optimal TSP tour is generated and then converted into TSP-D by local searches and (ii) the GRASP procedure, which optimally splits the TSP tour into TSP-D solutions. A new formulation of the TSP-D problem was presented by \cite{marinelli2018route} where the truck can also pick up a drone on a route arc (en route) rather than just at a customer location and heuristics to solve this were presented. \cite{tong2022optimal} proposed using variable neighbourhood tabu search to solve TSP-D. For FSTSP, \cite{ponza2016optimization} presented a heuristic approach using simulated annealing and \cite{de2020variable} used a hybrid heuristic where the initial solution is created from the optimal TSP solution reached by a TSP solver and then an implementation of general variable neighbourhood search is used to obtain the truck and drone delivery tours.

\subsection{Metaheuristics}
Metaheuristics have also been investigated for solving both TSP-D and FSTSP. For TSP-D, \cite{almuhaideb2021optimization} proposed using a greedy, randomised adaptive search procedure and a self-adaptive neighbourhood selection scheme. \cite{ha2020hybrid} proposed a hybrid genetic search algorithm with dynamic population management and adaptive diversity control. \cite{gunay2023hybrid} proposed a hybrid metaheuristic, combining a genetic algorithm with ant colony optimisation. For FSTSP, \cite{kuroswiski2023hybrid} proposed a hybrid genetic algorithm and \cite{peng2022optimization} proposed a multi-path GA, which calculates the minimum number of drones for delivery tasks under limited time conditions.

\section{Problem Description}
\label{problem}
FSTSP can be defined as a directed graph $G = (V, A)$, where $V$ is a set of nodes,  $V = \{1, 2, ..., n\}$, and $A$ is a set of arcs, defining the directed vehicle routes between nodes in the tour. The first node in the tour is always the depot, which cannot change, with customer nodes $C \subset V$,  $C= \{2, ..., n\}$. Each customer node is required to be visited exactly once by either the truck or drone. The depot node defines the start and end of each tour. The objective of the FSTSP problem is to minimise the total time required for the truck and drone to visit all customer locations and return to the depot, defined as the makespan. Given two nodes $i, j \in V$, we can denote $d_{T_{ij}}$ as the distance the truck has to drive from $i$ to $j$ and $d_{D_{ij}}$ as the distance the drone has to fly from $i$ to $j$. We can also denote the average speed of the truck and drone as $v_T$ and $v_D$ respectively. Therefore, given both the distance and time, we can calculate the amount of time required for the truck and drone to travel from $i$ to $j$, denoted as $t_{T_{ij}}$ and $t_{D_{ij}}$ respectively. The drone can launch from the truck at any location and then must visit a customer location and return to the truck at another customer location before the drone's endurance is exceeded. The process of the drone returning to the truck can require synchronisation where either vehicle arrives at the customer location before another. In this paper three node types are used, each representing which vehicle serves the location:
\begin{description}
    \item [Combined Node] A customer location served by the truck with the drone.
    \item [Drone Node] A customer location served by only a drone.
    \item [Truck-only Node] A customer location served only by the truck.
\end{description}

\subsection{Assumptions}
The following conditions are assumed:
\begin{itemize}
    \item Only one drone can be used at a time.
    \item Drones have unit capacity and have to return to the truck after each delivery. However, not immediately.
    \item The truck and drone must collectively visit each location exactly once, excluding the depot.
    \item Drones can only be deployed from a customer location and can only meet back with the truck at a different customer location.
    \item Drone endurance is infinite.
    \item Loading and drop-off time is negligible and can be ignored.
    \item The time to swap the drone's batteries is negligible and can be ignored.
\end{itemize}

\section{Proposed Self-adaptive Genetic Algorithm for FSTSP}
\label{proposed algorithm}
In this section, we describe the proposed steady-state genetic algorithm that is paired with a memeplex containing eight meme options, representing crossover and mutation operators, and the probabilities for each operator to be applied. The self-adaptive nature of this algorithm derives from its process of co-evolving the memeplex alongside the population each generation. Subsequently, adapting how the population is evolved each generation. The algorithm determines where the drone leaves the truck, which node it visits and at which node it rejoins with the truck. To begin with, Concord \cite{applegate2006concorde}, a TSP solver, is applied to the problem instance, generating the optimal TSP tour. Each node is initialised as a combined node. Then the population is generated; in this process, certain nodes in each new individual are converted into drone nodes. This process is explained in detail in section \ref{Population}. Then for $numGenerations$, two parents are selected for crossover by applying tournament selection twice. Once the parents have been selected, the offspring are produced by either applying the crossover operator present in the parent, with the lowest fitness, memeplex or simply inheriting the corresponding parent's chromosome. Once the offspring have been generated, mutation to both their tour and node types may be applied. The problem-tailored mutation operators are also taken from the memeplex of the parent with the lowest fitness. Following mutation of the offspring, mutation may be applied to the memeplex, inherited from the parent with the lowest fitness. Finally, an elitism-based population replacement strategy is applied to produce the next generation's population.

\begin{algorithm}
\caption{The Self-adaptive Genetic Algorithm}
    \begin{algorithmic}[1]
        \State $ tspTour \gets GetOptimalTspTour()$
        \State $ population \gets InitialisePopulation(tspTour)$
        \State $ i \gets 0$
        \While { $i < numGenerations$}
            \State $parent1 \gets SelectParent(population)$
            \State $parent2 \gets SelectParent(population)$
            \State $offspring1, offspring2 \gets ApplyCrossover(parent1, parent2)$
            \State $ ApplyMutation(offspring1, offspring2)$
            \State $ MutateMemeplex(offspring1, offspring2)$
            \State $ population \rightarrow ReplacePopulation(parent1, parent2, offspring1, offspring2)$
            \State $i \gets i + 1$
        \EndWhile
    \end{algorithmic}
    \label{Self-adaptive GA algorithm}
\end{algorithm}

\subsection{Solution Representation}

\begin{figure}[H]
    \centering
    \includegraphics[width=4in]{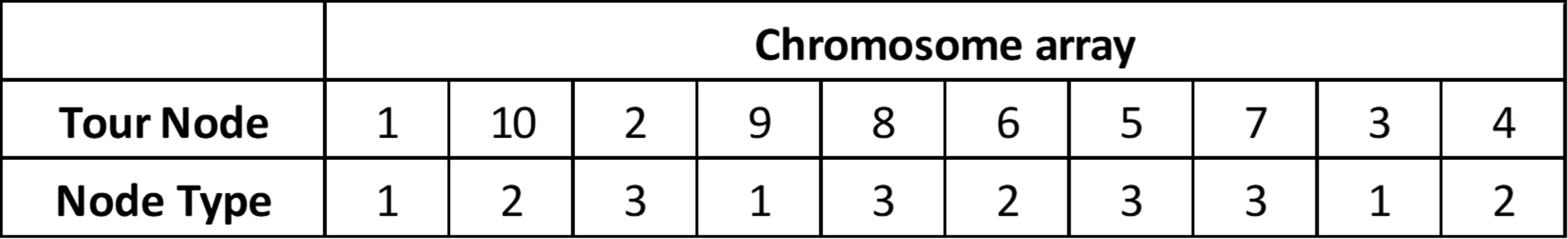}
    \caption{Corresponding chromosome array for the "singlecenter-51-n10" problem instance}
    \label{fig:chromosome array}
\end{figure}

The chromosome in each candidate solution is represented using an array of length N, where N is the number of nodes in the problem instance, including the depot. Each index in the array is used to represent a node in the candidate solution's tour. Hence, each index stores both the position of the node in the tour and the node's type. The first index is reserved for the depot node and therefore, is set as a combined node and cannot be modified. An example of an FSTSP tour for the "singlecenter-51-n10" problem instance provided by \cite{bouman2018instances} is illustrated in Figure \ref{fig:FSTSP tour} along with the corresponding chromosome array used to represent the solution in Figure \ref{fig:chromosome array}: 

\begin{figure}[H]
    \centering
    \includegraphics[width=5in]{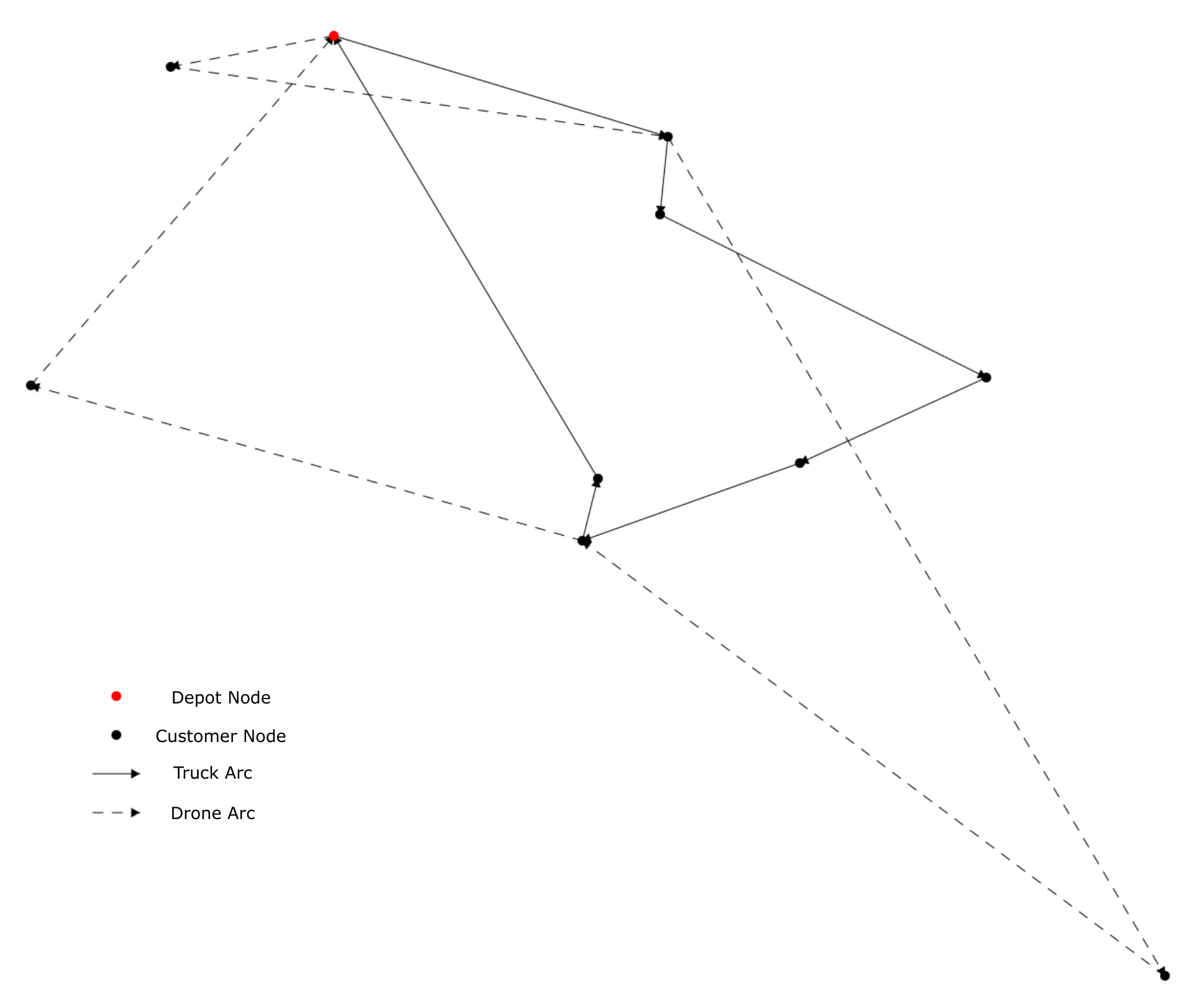}
    \caption{FSTSP tour of the "singlecenter-51-n10" problem instance}
    \label{fig:FSTSP tour}
\end{figure}

Each candidate solution is paired with a memeplex. The memeplex has eight options, each representing either an operator, mutation or crossover, or a probability corresponding to how often a specific type of operator should be applied. The meme options included are crossover operator (C), crossover probability (CP), combined mutation operator (CM), drone mutation operator (DM), truck mutation operator (TM), type mutation probability (TP), tour mutation operator (TOM), tour mutation probability (TOP). Figure \ref{fig:solution and memeplex} illustrates how each candidate is represented:

\begin{figure}[H]
    \centering
    \includegraphics[width=6.5in]{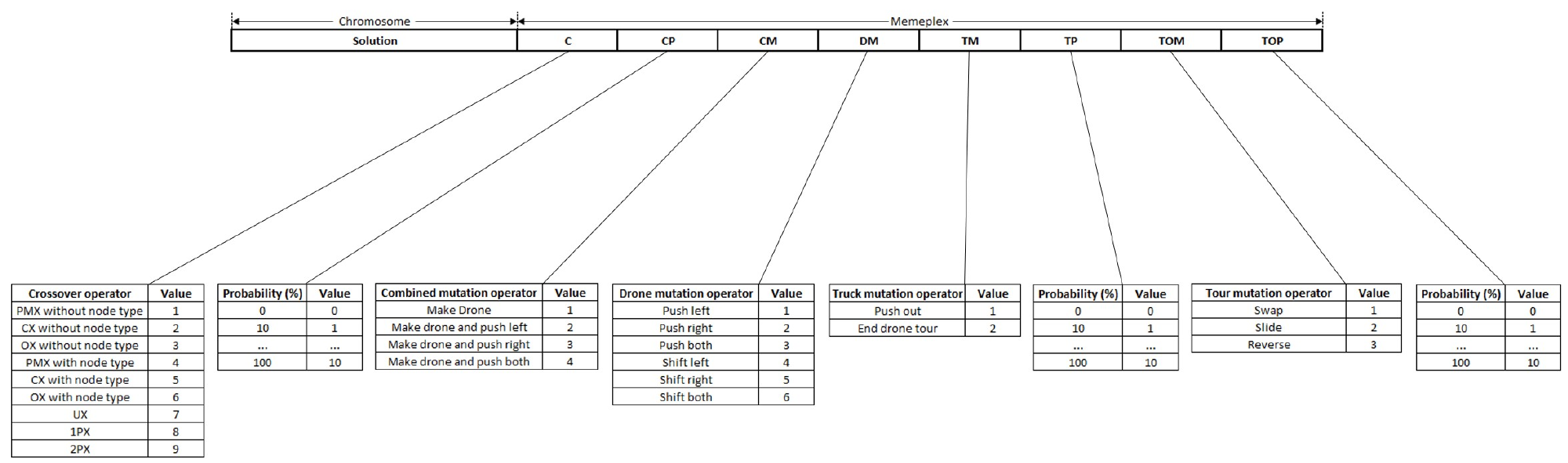}
    \caption{Solution and memeplex representation}
    \label{fig:solution and memeplex}
\end{figure}

\subsection{Objective Evaluation}

To save computation time and avoid recalculating the Euclidean distance between each pair of nodes every time an individual's fitness needs to be calculated, a distance matrix $d$ is constructed. Where $d_{ij}$ is the Euclidean distance between nodes $i$ and $j$, calculated simply using the Pythagorean theorem: 

\begin{equation}
    d_{ij} = \sqrt{(x_i - x_j)^2 + (y_i - y_j)^2} 
    \label{eq: }
\end{equation}

We denote $\alpha$ as the average speed of the drone relative to the average speed of the truck. In this paper $\alpha = 2$, therefore on average the drone's speed is twice that of the truck. This choice was made considering that the drone does not experience the same time delays as the truck does for travelling on the road. For example, the drone cannot get stuck in traffic, which would decrease its average speed. Hence, we can assume $v_T = 1$ and $v_D = \alpha$, giving the following formulas for travel time between nodes $i$ and $j$ for each vehicle:

\begin{equation}
    \begin{split}
        t_{T_{ij}} = d_{ij} \\\\
        t_{D_{ij}} = \frac{d_{ij}}{\alpha}
    \end{split}
\end{equation}

The makespan of a tour $T$ can be characterised by a sequence of subtour pairs, $S_T$ and $S_D$ representing the truck and drone tour respectively. Each subtour commences at a combined node and terminates at the rendezvous node, which must also be a combined node. To calculate the time required for a subtour pair $t_S$, the following formula is used:

$$
t_S(S) = \begin{cases}
    max(t_{S_{T}}, t_{S_{D}}), &  \quad \text{if } S \text{ contains a drone node} \\
    t_{S_{T}}, & \quad \text{if } S \text{ does not contain a drone node}
\end{cases}
$$

We can then calculate the makespan of a tour $T$ by computing the total time required for all subtour pairs in the tour, illustrated in the formula below:

\begin{equation}
    t_T = \sum_{s \in T} t_S(s)
\end{equation}

\subsection{Repair Operator}

A repair operator is required to legalise infeasible solutions produced by the crossover and mutation operators implemented in this algorithm. Two types of infeasible solutions can be encountered:
\begin{itemize}
    \item Two or more connected drone tours
    \item Disconnected truck-only nodes
\end{itemize}

In the first example, the repair operator begins by identifying multiple connected drone tours. It then calculates the midpoints between the drone nodes within these connected tours. These midpoint nodes are subsequently converted into combined nodes, serving as rendezvous nodes for the tours. In the second example, the repair operator identifies disconnected truck-only nodes and converts them into combined nodes, legalising the solution. Algorithm \ref{repair operator} illustrates this process.

\begin{algorithm}[H]
\caption{Repair operator}
    \begin{algorithmic}[1]
    \State $ i \gets 1$
    \While{$ i < populationSize $}
        \State $ currentNode \gets individual[i]$
        \If{ $ currentNode.type = 2$}
            \State $ nextDroneNode \gets FindNextDrone(currentNode)$
            \If{$ IsConnected(currentNode, nextDroneNode)$}
                \State $ midpoint \gets CalcMidpoint(currentNode, nextDroneNode)$
                \State $ midpoint.type \gets 1$
            \EndIf
        \ElsIf{$ currentNode.type = 3$ and $IsDisconnected(currentNode)$}
            \State $ currentNode.type \gets 1$
        \EndIf
        \State $i \gets i + 1$
    \EndWhile    
    \end{algorithmic}
    \label{repair operator}
\end{algorithm}

\subsection{Building the Initial Population}
\label{Population}

To generate the initial population, we assign a score to each node within the initial tour. This score is determined by calculating the makespan savings that would result from transforming the node into a drone node. We use the following formula to calculate the savings $s$, where $j$ denotes the current node, $i$ is the preceding node and $k$ is the succeeding node.

\begin{equation}
    s = max(t_{T_{ij}} + t_{T_{jk}} - max(t_{D_{ij}} + t_{D_{jk}}, t_{T_{ik}}),  1)
\end{equation}

The formula incorporates the maximum of the saving value and $1$ to ensure that when the scores are used in roulette wheel selection, there remains a possibility for every node to be chosen even if it doesn't offer an immediate saving to the tour's makespan. This adjustment was made to promote a wider range of solutions in the initial population and enable exploration of drone nodes that may not yield immediate but future makespan savings.

\begin{algorithm}[H]
\caption{Calculating node scores}
    \begin{algorithmic}[1]
        \State INITIALISE $scores$
        \State $ scores[0] \gets 0$
        \State $ j \gets 1$
        \While{$ j < numNodes$}
            \State $ i \gets j - 1$
            \If{$ j = (numNodes - 1)$}
                \State $ k \gets 0$
            \Else
                \State $ k \gets j + 1$
            \EndIf
            \State $ scores[j] \gets max(tTij + tTjk - max(tDij + tDjk, tTik), 1)$
            \State $j \gets j + 1$
        \EndWhile
        \State \Return $ scores$
    \end{algorithmic}
\end{algorithm}

Once the scores array has been calculated, the population is generated. This process involves applying roulette wheel selection to select and convert certain nodes into drone nodes using the scores array to associate a probability of selection to each node. 
Given a node $i$, the probability of it being selected is denoted as $p_i$. If $s_i$ is the score of node $i$, the probability of it being selected is: 

\begin{equation}
    p_i = \frac{s_i}{\sum_{j=1}^{N}s_j}
\end{equation}

\begin{figure}[H]
    \centering
    \includegraphics[width = 5.5in]{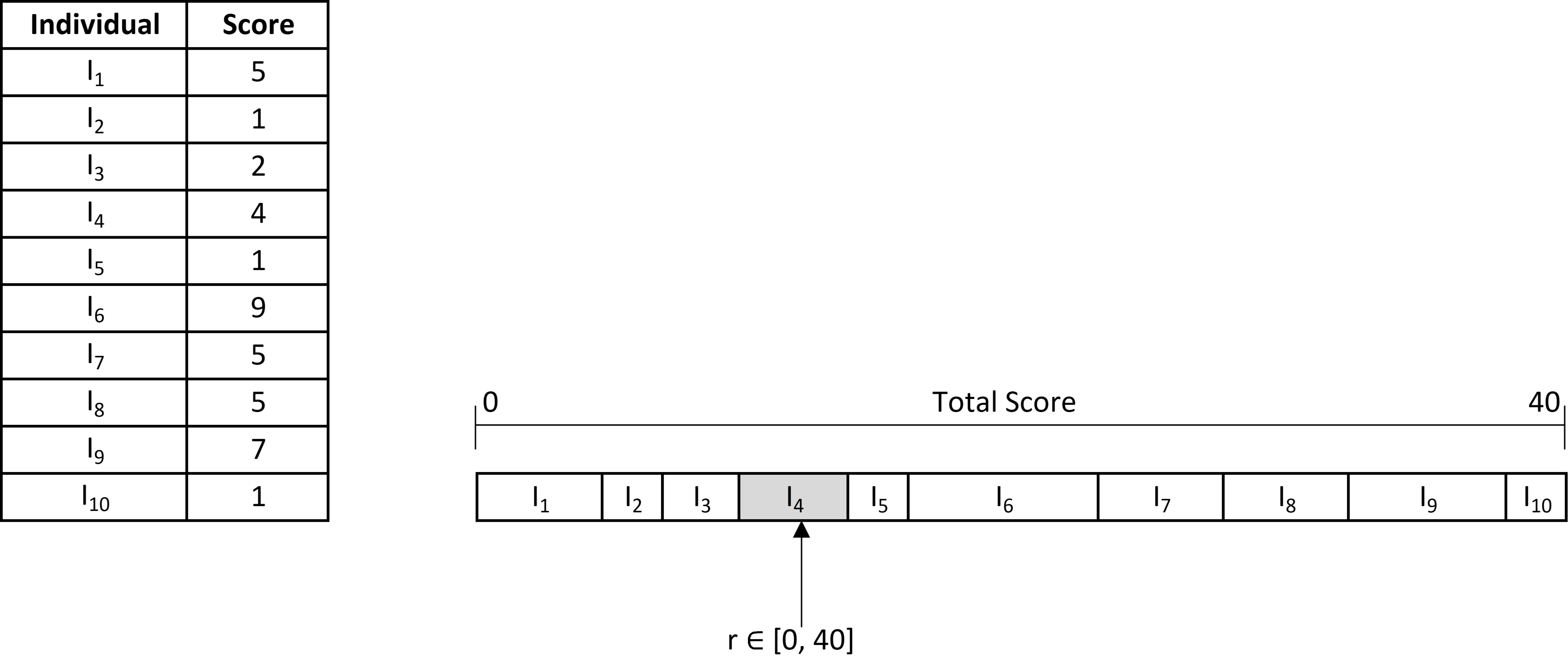}
    \caption{An example of roulette wheel selection}
    \label{fig:roulette wheel selection}
\end{figure}

Where $N$ denotes the number of nodes in the tour.
\\An example of applying roulette wheel selection is shown in Figure \ref{fig:roulette wheel selection}. Once, a node has been selected, that node in the individual is converted into a drone node and removed from the current score array so that it cannot be selected again. This process is repeated $initialDrone\% \times N$ times for each individual. Then, the entire process is repeated $numIndividuals$ times to generate the entire population.

\begin{algorithm}[H]
\caption{Generating initial population}
    \begin{algorithmic}[1]
        \State $ scores \gets GetDroneScores()$
        \State $ i \gets 0$
        \While{$ i < numIndividuals$}
            \State INITIALISE $newIndividual$
            \State $ currentScores \gets scores$
            \State $ j \gets 0$
            \While{$ j < (numNodes / numInitialDrones)$}
                \State $selectedNode \gets ApplyRouletteWheel(currentScores)$
                \State $currentScores[selectedNode] \gets 0$
                \State $newIndividual[selectedNode].type \gets 2$
                \State $j \gets j + 1$
            \EndWhile
            \State $ EvaluateFitness(newIndividual)$
            \State $ AddToPopulation(newIndividual)$
            \State $i \gets i + 1$
        \EndWhile
    \end{algorithmic}
\end{algorithm}

\subsection{Parent Selection}
\begin{figure}[H]
    \centering
    \includegraphics[width = 5.5in]{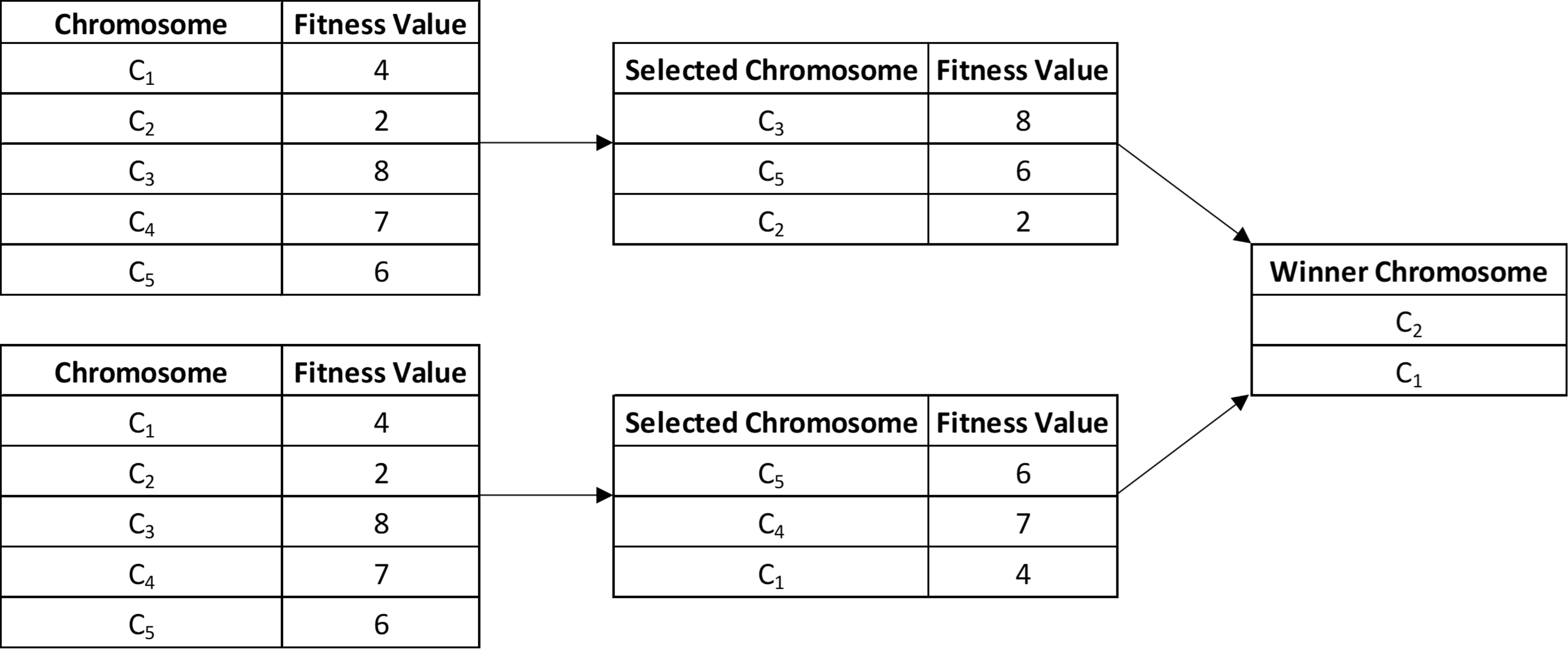}
    \caption{An example of tournament selection}
    \label{fig:tournament selection}
\end{figure}

To select each generation's parents, tournament selection is used. In this process, $t$-individuals in the population are randomly chosen. The fitness of each individual's chromosome is evaluated, and the individual with the lowest fitness value is selected. This process is repeated twice to generate two parents for either crossover or simple inheritance, where the children directly inherit the corresponding parent's chromosome. An example of this process is illustrated in Figure \ref{fig:tournament selection}, with $t = 3$.

The probability of selecting an individual $i$ is denoted as $p_i$. Under the assumption that the individuals in the population are sorted according to their fitness value $f(i_1) \leq f(i_2) \leq ... \leq f(i_n)$, where $f(i)$ is the fitness of individual $i$, the probability of it being selected is:

\begin{equation}
    p_i = N^{-t}((N - i + 1)^{t} - (N - i)^{t})
\end{equation}

Where $N$ denotes the number of individuals in the population and $t$ denotes the tour size.

\subsection{Crossover}
Once the parents have been selected, the crossover operator $C$, specified in the memeplex of the parent with the lowest fitness, may be applied. The probability of the operator being applied is also specified in the same parent's memeplex, denoted by $CP$. For example, if $CP = 8$, the chance of applying crossover to generate each offspring's chromosome is 80\%. If the crossover operator is not used, the offspring directly inherit their chromosome from the corresponding parent. For example, offspring 1 would inherit their chromosome from parent 1. In total, there are 6 different crossover operators used in this algorithm, presented by \cite{kora2017crossover}. However, partially mapped crossover (PMX), cycle crossover (CX) and order crossover (OX) are used to both crossover the tour sequence and node type and just the tour sequence. Hence, creating 9 unique operators:

\begin{description}
    \item[Partially Mapped Crossover (PMX)] Each parent's chromosome is cut into three substrings based on two randomly generated points. Then each offspring inherits the gene substring located between the two cut points from the opposite parent. A mapping relationship is then determined, mapping the alleles at each locus in the substring between the corresponding parent and offspring. Each offspring's chromosome is then legalised by filling the remaining empty genes using the same locus in the corresponding parent's chromosome. The mapping relationship is then used if the parent's gene contains an allele already present in the offspring.
    \item [Cycle Crossover (CX)] Cycles are identified between the parent's chromosomes commencing at a random locus. Using this cycle, the offspring are generated. For instance, offspring 1 will inherit each gene from parent 1's chromosome that is in the cycle at the same locus. Then each offspring's chromosome is legalised by filling the remaining empty genes using the same locus in the opposite parent's chromosome.
    \item [Order Crossover (OX)] Similarly to PMX, each parent's chromosome is cut into three substrings based on two randomly generated points. Then each offspring inherits the gene substring located between the two cut points from the corresponding parent. Then to legalise the offspring's chromosomes, genes are directly inherited from the opposite parent's chromosome starting at the locus directly after the second cut point and wrapping back to the first locus when the end of the chromosome is reached. Genes from the parent are skipped if the gene's allele is already present in the offspring. 
    \item [Uniform Crossover (UX)] For each locus in each offspring's chromosome, there is an equal chance that either the allele is inherited from the same locus in the corresponding parent's chromosome or that the allele is inherited from the same locus in the opposite parent's chromosome.
    \item [1-point Crossover (1px)] One cut point is randomly generated. For each locus before the cut point, the offspring inherits the allele from the corresponding parent's chromosome. Then, for each locus after the cut point, the offspring inherits the allele from the opposite parent's chromosome.
    \item [2-point Crossover (2px)] Two cut points are randomly generated. For each locus before the first cut point and after the second, the offspring inherits the allele from the opposite parent's chromosome. Then, for each locus between the two cut points, the offspring inherits the allele from the corresponding parent's chromosome.
\end{description}

\subsection{Mutation}
After the chromosome for each offspring has been created, two types of mutation are applied. By dividing the mutation process into two distinct types, it enables the opportunity for more significant changes to the offspring's chromosome. This is because both the tour sequence and node types can be modified in a single generation. The tour sequence operator is specified by the tour mutation operator (TOM) meme option present in the memeplex of the parent with the lowest fitness. Mutation of the tour sequence takes place first and three specific operators are used. The probability of tour sequence mutation is dictated by the tour mutation probability (TOP) meme option. The following tour sequence mutation operators are used:

\textbf{Tour sequence mutation operators:}
\begin{description}
    \item [Swap mutation] Two nodes are selected at random and their order in the tour sequence is swapped.
    \item [Slide mutation] Two nodes are selected at random. Then for each node between the two selected, the node is shifted to the left. To legalise the new tour, the last node is set to the original first node.
    \item [Reverse mutation] Two nodes are selected at random. Then for each node between the two selected, the order of the nodes is reversed. 
\end{description}

Within the node type mutation, 12 different problem-tailored operators were used, split into three subcategories specific to the type of node they are applicable to. The operators: Make fly, push left and push right were proposed by \cite{almuhaideb2021optimization} for TSP-D. To apply node type mutation, a node in the tour is randomly selected. Then, corresponding to the node's type, the corresponding mutation operator is taken from the memeplex of the parent with the lowest fitness and then applied to the node. For example, if the selected node is truck-only, the operator is taken from the truck mutation (TM) meme option. The probability of type mutation being applied is dependent on the type mutation probability (TP) meme option.

\textbf{Drone mutation operators:}
\begin{description}
    \item [Push left] The node to the left of the drone node, if it isn't the depot, is converted into a truck-only node.
    \item [Push right] The node to the right of the drone node, if it exists, is converted into a truck-only node.
    \item [Push both] The neighbouring nodes of the drone node, if it's possible, are converted into truck-only nodes.
    \item [Shift left] The node to the left of the drone node, if it isn't the depot, is converted into a drone node. Then the original drone node is converted into either a truck-only node or a combined node. This is dependent on whether the drone tour continued after the original drone node. 
    \item [Shift right] The node to the right of the drone node, if it exists, is converted into a drone node and the original drone node is converted into either a truck-only node or a combined node. This is dependent on whether the drone tour commenced before the original drone node. 
    \item [Shift both] Both the shift left operator and the shift right operator are applied.
\end{description}

\textbf{Combined mutation operators:}
\begin{description}
    \item [Make fly] The node is converted into a drone node. 
    \item [Make fly and push left] The node is converted into a drone node and the push left operator is applied.
    \item [Make fly and push right] The node is converted into a drone node and the push right operator is applied.
    \item [Make fly and push both] The node is converted into a drone node and both the push left and push right operators are applied.
\end{description}

\textbf{Truck-only mutation operators:}
\begin{description}
    \item [Push out] The drone tour, if possible, is extended in accordance with the position of the current truck-only node relative to the drone node.
    \item [End drone tour] The current truck-only node is converted into a combined node. Subsequently, any nodes that are now unconnected are also converted into combined nodes.
\end{description}

Once mutation has been applied to each offspring's chromosome, the memeplex of each offspring is also mutated. This is dependent on the innovation rate hyper-parameter. For example, to determine if a meme is mutated, a random value is generated, $r \in [0, 10]$. If $r < innovationRate$, then the meme option is set to a new random value. Hence, the innovation rate can be used to adjust the probability and frequency at which each meme option in the memeplex is mutated.

\subsection{Population Replacement}
Initially, each offspring has their fitness evaluated. Then, out of the parents and offspring, the two individuals with the lowest fitness are selected to replace the parents in the next generation. This elitist population replacement mechanism was chosen to maintain diversity, especially of lower quality solutions while intending to reduce the number of identical, both genotypic and phenotypic, solutions occupying the population. An example of this population replacement process is shown in Figure \ref{fig:population replacement}:

\begin{figure}[H]
    \centering
    \includegraphics[width = 6in]{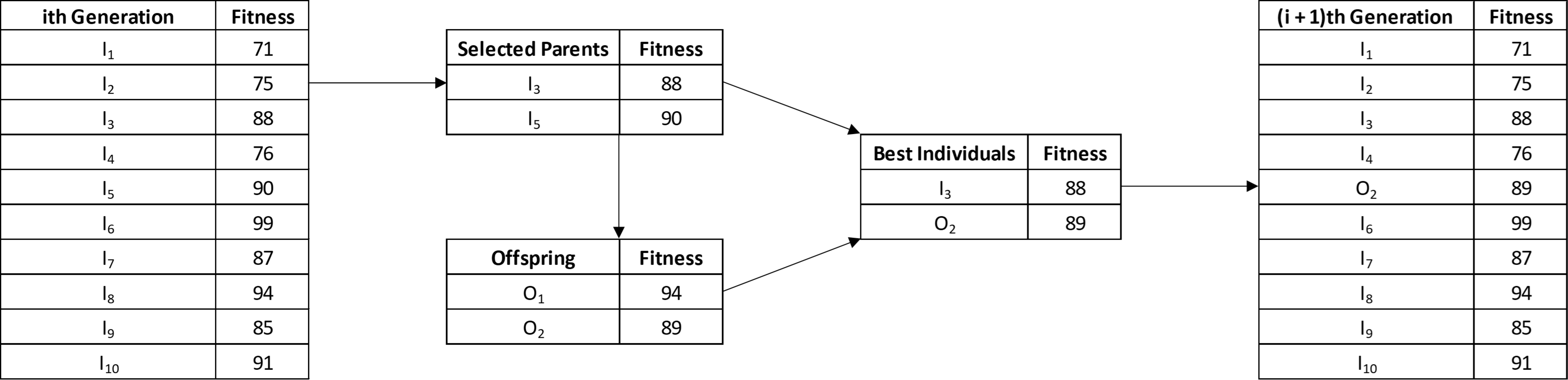}
    \caption{An example of population replacement}
    \label{fig:population replacement}
\end{figure} 

\section{Experimental Results}
\label{results}
The following section presents the computational experiments used to set the hyper-parameter values and evaluate the performance of the proposed algorithm. The algorithm was implemented in Visual Studio C++ and the experiments were run on an Intel Core i7-8700T processor at 2.4GHz with 32GB of RAM. 

\subsection{Hyper-parameter setting}
To determine this algorithm’s hyper-parameter values, including innovation rate, initial drone percentage, population size (in thousands), and tour size, we employed a Taguchi design. This approach involved testing 16 different combinations of hyper-parameter levels (L16) to identify the best values for each hyper-parameter independently. To ensure the computation time remains reasonably low, the number of generations was set to one million. These experiments used the problem instance "singlecenter-63-n20" provided by \cite{bouman2018instances}, defining a 20-node graph. Table \ref{tab:Parameter levels} presents the hyper-parameters used at each level for the Taguchi design and table \ref{tab:parameter experiments} illustrates the average fitness achieved after 30 trials on each experiment.

\begin{table}[H]
 \caption{Hyper-parameter levels for Taguchi design}
   \centering
  \begin{tabular}{lcccc}
    \toprule
    Hyper-parameter & Level 1 & Level 2 & Level 3 & Level 4\\
    \midrule
    Innovation rate & 6 & 7 & 8 & 9 \\
    Initial drone \% & 2 & 5 & 8 & 11 \\
    Population size & 20 & 30 & 40 & 50 \\
    Tour size & 5 & 20 & 35 & 50 \\
    \bottomrule
  \end{tabular}
  \label{tab:Parameter levels}
\end{table}

\begin{table}[H]
 \caption{Taguchi design experiments}
   \centering
  \begin{tabular}{cccccc}
    \toprule
    Config Number & Innovation rate & Initial drone \% & Population size (Thousands) & Tour size & Average fitness\\
    \midrule
    1 & 6  & 2  & 20 & 5 & 452.63 \\
    2 & 6  & 5  & 30 & 20 & 463.24 \\
    3 & 6  & 8  & 40 & 35 & 465.94 \\
    4 & 6  & 11 & 50 & 50& 469.65 \\
    5 & 7  & 2 & 30 & 35& 461.56 \\
    6 & 7  & 5 & 20 & 50 & 470.03 \\
    7 & 7  & 8 & 50 & 5 & 448.68  \\
    8 & 7  & 11 & 40 & 20 & 465.61 \\
    9 & 8  & 2 & 40 & 50 & 466.12 \\
    10 & 8  & 5 & 50 & 35 & 461.12 \\
    11 & 8  & 8 & 20 & 20& 464.75 \\
    12 & 8  & 11 & 30 & 5& 454.75 \\
    13 & 9  & 2 & 50 & 20 & 456.93 \\
    14 & 9  & 5 & 40 & 5 & 448.91 \\
    15 & 9  & 8 & 30 & 50 & 471.91 \\
    16 & 9  & 11 & 20 & 35 & 470.67 \\
    \bottomrule
  \end{tabular}
  \label{tab:parameter experiments}
\end{table}

\begin{figure}[H]
    \centering
    \includegraphics[width = 6in]{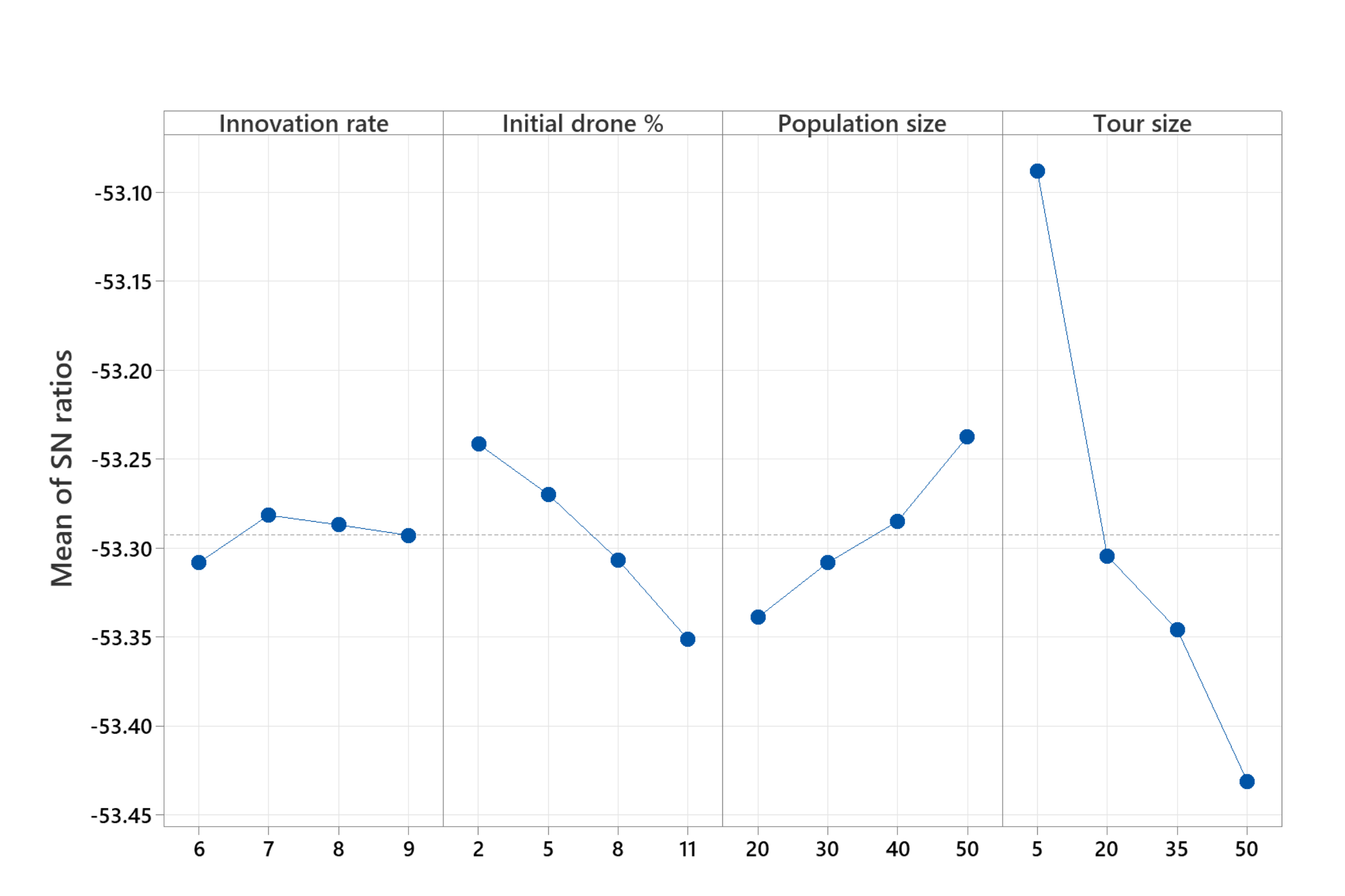}
    \caption{Main effects plot for SN ratios}
    \label{fig:sn ratios plot}
\end{figure}

Subsequently, the results from the Taguchi design were assessed by examining the main effects for means and signal-to-noise (SN) ratios. In the main effects for means, the lowest value indicates the best level for each hyper-parameter. Whereas, in the main effects for SN ratios, the highest value indicates the best level for the hyper-parameter. The Figures \ref{fig:means plot} and \ref{fig:sn ratios plot} illustrate the main effects and SN ratios at each respective level:

\begin{figure}[H]
    \centering
    \includegraphics[width = 6in]{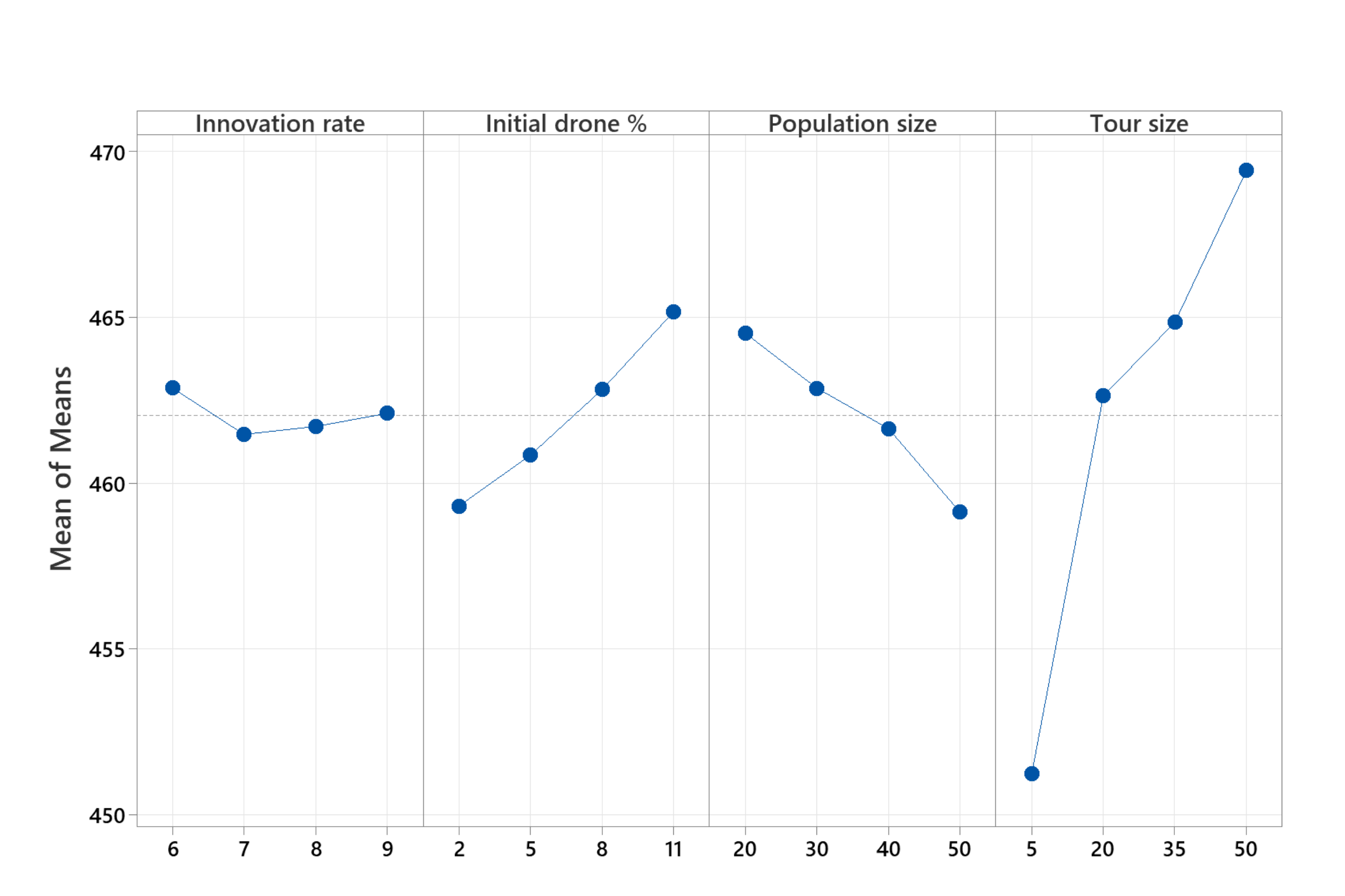}
    \caption{Main effects plot for means}
    \label{fig:means plot}
\end{figure} 

\begin{figure}[H]
    \centering
    \includegraphics[width = 5.5in]{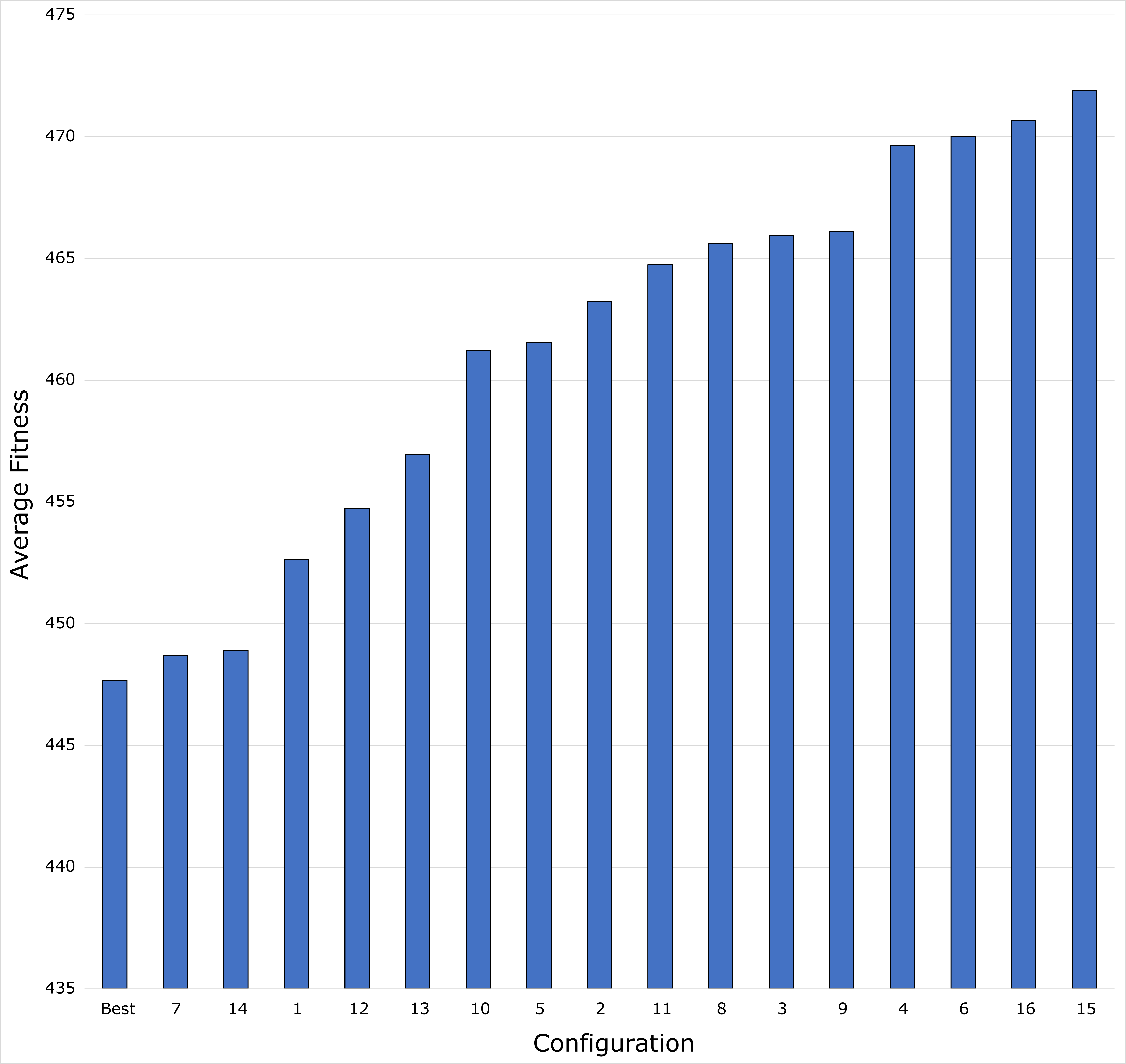}
    \caption{Confirmation testing chart}
    \label{fig:confirmation chart}
\end{figure}

Following these experiments, the hyper-parameter values were established based on the best-performing levels, illustrated in Table \ref{tab:selected parameters}. To confirm and validate that the selected configuration produces the lowest average fitness, a validation test was run, illustrated in Figure \ref{fig:confirmation chart} with the selected configuration labelled "best".

\begin{table}[H]
 \caption{Selected hyper-parameter values}
   \centering
  \begin{tabular}{lc}
    \toprule
    Hyper-parameter & Value \\
    \midrule
    Innovation rate & 7 \\
    Initial drone \% & 2 \\
    Population size & 50 \\
    Tour size & 5 \\
    \bottomrule
  \end{tabular}
  \label{tab:selected parameters}
\end{table}

Figure \ref{fig:confirmation chart} clearly illustrates that the selected hyper-parameter values are the best configuration for this algorithm as it results in the lowest average fitness compared to every other configuration tested. Consequently, this configuration of hyper-parameter values has been used for the remaining tests in this paper.

\subsection{Results on Smaller-Sized Instances}
To commence testing, this algorithm was evaluated using smaller-sized single-centre problem instances provided by \cite{bouman2018instances}. Five different sizes were used, $n = \{5, ..., 9\}$, where $n$ is the number of locations in the problem instance. In table \ref{tab:smaller sized results}, the column labelled "n" presents the number of locations. These smaller-sized problem instances have been solved using the ILP formulation of the problem and hence, their optimal solutions are included by \cite{bouman2018instances}. In table \ref{tab:smaller sized results} below, the optimal solution value is presented in the column labelled "IP". For each problem size, \cite{bouman2018instances} provides ten different instances and therefore in total, 50 different experiments were run. Each experiment was run for 30 trials with the best solution obtained presented in the column labelled "Self-Adaptive GA". The last column, labelled "GAP\%" illustrates the percentage difference between the optimal solution, provided by \cite{bouman2018instances}, and the best-found solution by this algorithm.

\begin{table}[H]
 \caption{Results on smaller-sized problem instances from \cite{bouman2018instances}}
   \centering
  \begin{tabular}{ccccc|ccccc}
    \toprule
    Instance & n & IP & Self-Adaptive GA & Gap\% & Instance & n & IP & Self-Adaptive GA & Gap\% \\
    \midrule
    1 & 5 & 154.26 & 156.76 & 1.62 & 26 & 7 & 98.71 & 98.77 & 0.06 \\
    2 & 5 & 140.54 & 140.54 & 0 & 27 & 7 & 177.86 & 177.86 & 0 \\
    3 & 5 & 52.67 & 52.92 & 0.47 & 28 & 7 & 169.77 & 169.77 & 0 \\
    4 & 5 & 108.94 & 108.94 & 0 & 29 & 7 & 193.34 & 193.34 & 0 \\
    5 & 5 & 122.15 & 122.15 & 0 & 30 & 7 & 177.10 & 177.10 & 0 \\
    6 & 5 & 162.22 & 162.22 & 0 & 31 & 8 & 155.20 & 155.20 & 0 \\
    7 & 5 & 133.95 & 133.95 & 0 & 32 & 8 & 107.20 & 107.20 & 0 \\
    8 & 5 & 81.50 & 81.50 & 0 & 33 & 8 & 172.85 & 175.50 & 1.53 \\
    9 & 5 & 143.15 & 143.15 & 0 & 34 & 8 & 226.91 & 226.91 & 0 \\
    10 & 5 & 140.40 & 140.40 & 0 & 35 & 8 & 188.46 & 188.46 & 0 \\
    11 & 6 & 128.01 & 128.01 & 0 & 36 & 8 & 181.37 & 181.37 & 0 \\
    12 & 6 & 125.94 & 125.94 & 0 & 37 & 8 & 134.62 & 134.62 & 0 \\
    13 & 6 & 215.91 & 216.11 & 0.09 & 38 & 8 & 286.71 & 286.71 & 0 \\
    14 & 6 & 119.24 & 119.24 & 0 & 39 & 8 & 181.07 & 181.07 & 0 \\
    15 & 6 & 169.62 & 169.62 & 0 & 40 & 8 & 214.90 & 214.90 & 0 \\
    16 & 6 & 117.84 & 117.84 & 0 & 41 & 9 & 116.93 & 116.93 & 0 \\
    17 & 6 & 263.03 & 263.03 & 0 & 42 & 9 & 316.25 & 316.25 & 0 \\
    18 & 6 & 258.65 & 258.65 & 0 & 43 & 9 & 226.28 & 243.01 & 7.39 \\
    19 & 6 & 188.80 & 188.80 & 0 & 44 & 9 & 228.28 & 228.28 & 0 \\
    20 & 6 & 122.17 & 122.17 & 0 & 45 & 9 & 279.66 & 279.66 & 0 \\
    21 & 7 & 208.34 & 209.60 & 0.60 & 46 & 9 & 214.02 & 216.19 & 1.01 \\
    22 & 7 & 178.38 & 178.38 & 0 & 47 & 9 & 277.73 & 277.73 & 0 \\
    23 & 7 & 116.24 & 116.24 & 0 & 48 & 9 & 200.95 & 200.95 & 0 \\
    24 & 7 & 152.59 & 152.59 & 0 & 49 & 9 & 349.49 & 349.49 & 0 \\
    25 & 7 & 130.50 & 130.50 & 0 & 50 & 9 & 196.84 & 196.84 & 0 \\
    \bottomrule
  \end{tabular}
  \label{tab:smaller sized results}
\end{table}

Assessing the performance of this algorithm, it was able to find the optimal solution on 42 out of the 50 problem instances tested. Moreover, the average percentage gap between the best solution found and the optimal solution was only 0.26\%. This demonstrates that this algorithm is highly effective at finding optimal solutions to smaller-sized problems and when it cannot, solutions that are very close in quality to the optimal are discovered. Comparing these results to a rival algorithm, this algorithm was able to find five more optimal solutions than \cite{gunay2023hybrid} and has a lower average gap \% to the optimal solutions, 0.26\% compared to 0.8\% by \cite{gunay2023hybrid}.

\subsection{Results on Large-Sized Instances}

To further the evaluation of this algorithm, larger-sized problem instances provided by \cite{bouman2018instances} were used, where $n = \{10, 20, 50, 75\}$. For each $n$-sized problem, ten different instances were provided, resulting in 40 different problem instances being tested against. For these larger-sized problems, \cite{bouman2018instances} have not provided any optimal solutions, hence it is beneficial to compare the results of this algorithm to those obtained by rival algorithms. LS \cite{agatz2018optimization}, HGVNS \cite{de2020variable}, and the two variants of the GRASP algorithm \cite{almuhaideb2021optimization} and \cite{gunay2023hybrid}'s GA-AS were used. For each $n$-sized problem, the mean of the best-found solutions for each algorithm is presented in Table \ref{tab:larger sized results}. Each problem instance was run for 30 trials.

\begin{table}[H]
    \caption{Results on larger size problem instances from \cite{bouman2018instances} and comparison with rivals \cite{agatz2018optimization, de2020variable, almuhaideb2021optimization, gunay2023hybrid}}
   \centering
  \begin{tabular}{ccccccc}
    \toprule
    n & LS \cite{agatz2018optimization} & HGVNS \cite{de2020variable} & GRASP-HCLS \cite{almuhaideb2021optimization} & GRASP-SA \cite{almuhaideb2021optimization} & GA-AS \cite{gunay2023hybrid} & Self-Adaptive GA \\
    \midrule
    10 & 278.22 & 291.36 & 287.13 & 295.49 & 263.50 & \textbf{261.39} \\
    20 & 384.87 & 364.08 & 416.47 & 428.09 & 399.73 & \textbf{359.26} \\
    50 & 554.58 & 593.54 & 617.42 & 723.88 & 649.61 & \textbf{527.49} \\
    75 & 741.38 & 754.43 & 861.21 & 1030.92 & 978.25 & \textbf{736.99} \\
    \bottomrule
  \end{tabular}
  \label{tab:larger sized results}
\end{table}

The proposed Self-Adaptive GA algorithm outperformed every rival algorithm on each $n$-sized problem and improved the best-found solutions. Specifically, the best-found solution for the $n = 50$ problem was improved by 5\%. Thus, demonstrating that this algorithm is also highly effective at solving larger-sized problem instances. 

\subsection{Computation Time}

Table \ref{tab:computation time} below describes the average computation of this algorithm on a single problem instance for each $n$-sized problem over 30 trials. These results demonstrate that this algorithm can find results for problem instances with $n < 10$ extremely quickly. Furthermore, this algorithm can find new best-known solutions to larger instances, $n > 9$, in very reasonable times. Specifically, compared to \cite{gunay2023hybrid} this algorithm's computation time is significantly lower on larger problems. Where $n = \{50, 75\}$, this algorithm outperforms \cite{gunay2023hybrid} by 48.53 seconds and 148.46 seconds respectively. However, this comparison is limited as both algorithms were tested on different systems which can affect computation time considerably.

\begin{table}[H]
 \caption{Average computation time on \cite{bouman2018instances} problem instances}
   \centering
  \begin{tabular}{cc|cc}
    \toprule
    n & Time (Seconds) & n & Time (Seconds) \\
    \midrule
    5 & 2.25 & 10 & 2.53 \\
    6 & 2.26 & 20 & 4.12 \\
    7 & 2.20 & 50 & 26.38 \\
    8 & 2.43 & 75 & 48.54 \\
    9 & 2.56 & \\
    \bottomrule
  \end{tabular}
  \label{tab:computation time}
\end{table}

\section{Conclusion}
\label{conclusion}

To conclude, this paper has introduced a novel and self-adaptive genetic algorithm (GA) tailored for solving the Flying Sidekick Travelling Salesman Problem (FSTSP). The literature in the field of metaheuristics for FSTSP has been relatively limited, with only two prior studies \cite{kuroswiski2023hybrid, peng2022optimization}, providing the motivation for this paper. The main contribution of this research lies in the development of a self-adaptive GA that demonstrates greater adaptability to the problem by incorporating a wide range of crossover operators, problem-specific mutation operators and utilising a memeplex to direct the usage of each operator over each generation. Moreover, this paper introduces a novel two-stage mutation process designed specifically to address the complexities of FSTSP, where both the tour sequence and node types can be modified in a single generation. The results of the experiments conducted showcased the effectiveness of the proposed self-adaptive GA in finding optimal solutions for smaller-sized problem instances. Additionally, the experiments also demonstrated this algorithm's strong capability to achieve novel best-known solutions for larger-sized problem instances, outperforming all rival algorithms. One of the restrictions made in this paper was limiting the number of drones that can fly simultaneously to one. Future studies could explore the enhancements in solution quality achieved by varying the number of drones.


\bibliographystyle{unsrt}  
\bibliography{references}

\end{document}